\documentclass[journal]{IEEEtran}

\usepackage{amsmath,amssymb,amsfonts}
\usepackage{graphicx}
\usepackage{textcomp}

\usepackage{amsthm}
\usepackage{xcolor}
\usepackage[figuresright]{rotating}

\usepackage{graphicx}
\graphicspath{{/}{fig/}}

\usepackage{array}
\usepackage{textcomp}
\usepackage{xcolor}
\usepackage{multirow}
\usepackage{booktabs}
\usepackage{enumitem}

\usepackage{mathtools}
\usepackage{breqn}
\usepackage{float}

\usepackage{caption}
\usepackage{subcaption}

\usepackage{mathtools}
\usepackage{float}
\usepackage{hyperref}

\usepackage{pgfplots}
\pgfplotsset{compat=1.7}
\usepgfplotslibrary{groupplots}

\usepackage{multirow,tabularx}
\usepackage{flushend}

\newlength\figureheight
\newlength\figurewidth
\title{\LARGE \bf
    Towards Closing the Sim-to-Real Gap in \\ Collaborative Multi-Robot Deep Reinforcement Learning
}

\author{
    \IEEEauthorblockN{
        Wenshuai Zhao\textsuperscript{1},
        Jorge Pe\~{n}a Queralta\textsuperscript{1},
        Li Qingqing\textsuperscript{1},
        Tomi Westerlund\textsuperscript{1}
    }\\[6pt]
    \IEEEauthorblockA{
        \textsuperscript{1} \href{https://tiers.utu.fi}{Turku Intelligent Embedded and Robotic Systems Lab, University of Turku, Finland} \\
        Emails: \textsuperscript{1}\{wezhao, jopequ, qingqli, tovewe\}@utu.fi
    }
}

\begin{document}

\maketitle

\begin{abstract}

    Current research directions in deep reinforcement learning include bridging the simulation-reality gap, improving sample efficiency of experiences in distributed multi-agent reinforcement learning, together with the development of robust methods against adversarial agents in distributed learning, among many others. In this work, we are particularly interested in analyzing how multi-agent reinforcement learning can bridge the gap to reality in distributed multi-robot systems where the operation of the different robots is not necessarily homogeneous. These variations can happen due to sensing mismatches, inherent errors in terms of calibration of the mechanical joints, or simple differences in accuracy. While our results are simulation-based, we introduce the effect of sensing, calibration, and accuracy mismatches in distributed reinforcement learning with proximal policy optimization (PPO). We discuss on how both the different types of perturbances and how the number of agents experiencing those perturbances affect the collaborative learning effort. The simulations are carried out using a Kuka arm model in the Bullet physics engine. This is, to the best of our knowledge, the first work exploring the limitations of PPO in multi-robot systems when considering that different robots might be exposed to different environments where their sensors or actuators have induced errors. With the conclusions of this work, we set the initial point for future work on designing and developing methods to achieve robust reinforcement learning on the presence of real-world perturbances that might differ within a multi-robot system.

\end{abstract}

\begin{IEEEkeywords}
    Reinforcement Learning; Multi-Robot Systems; Collaborative Learning; Deep RL; Adversarial RL; Sim-to-Real;
\end{IEEEkeywords}

\IEEEpeerreviewmaketitle

\section{Introduction}

Reinforcement learning (RL) algorithms for robotics and cyber-physical systems have seen an increasing adoption across multiple domains over the past decade~\cite{arulkumaran2017brief, nguyen2020deep}. Deep reinforcement learning (DRL) enables agents to be trained in realistic environments without the need for large amounts of data to be gathered and labeled a priori. Specifically, reinforcement learning has enjoyed significant success in robotic control tasks involving manipulation~\cite{mnih2016asynchronous, rajeswaran2017learning, matas2018sim}. Motivated by the way humans and animals learn, DRL algorithms work on a \textit{trial and error} basis, where an agent interacts with its environment and receives a reward based on its performance. When complex agents or environments are involved, the learning process can be relatively slow. This has motivated the design and development of multi-agent DRL algorithms. In this paper, we are interested in exploring some of the challenges remaining in multi-robot collaborative DRL.

\begin{figure}
    \centering
    \includegraphics[width=0.48\textwidth]{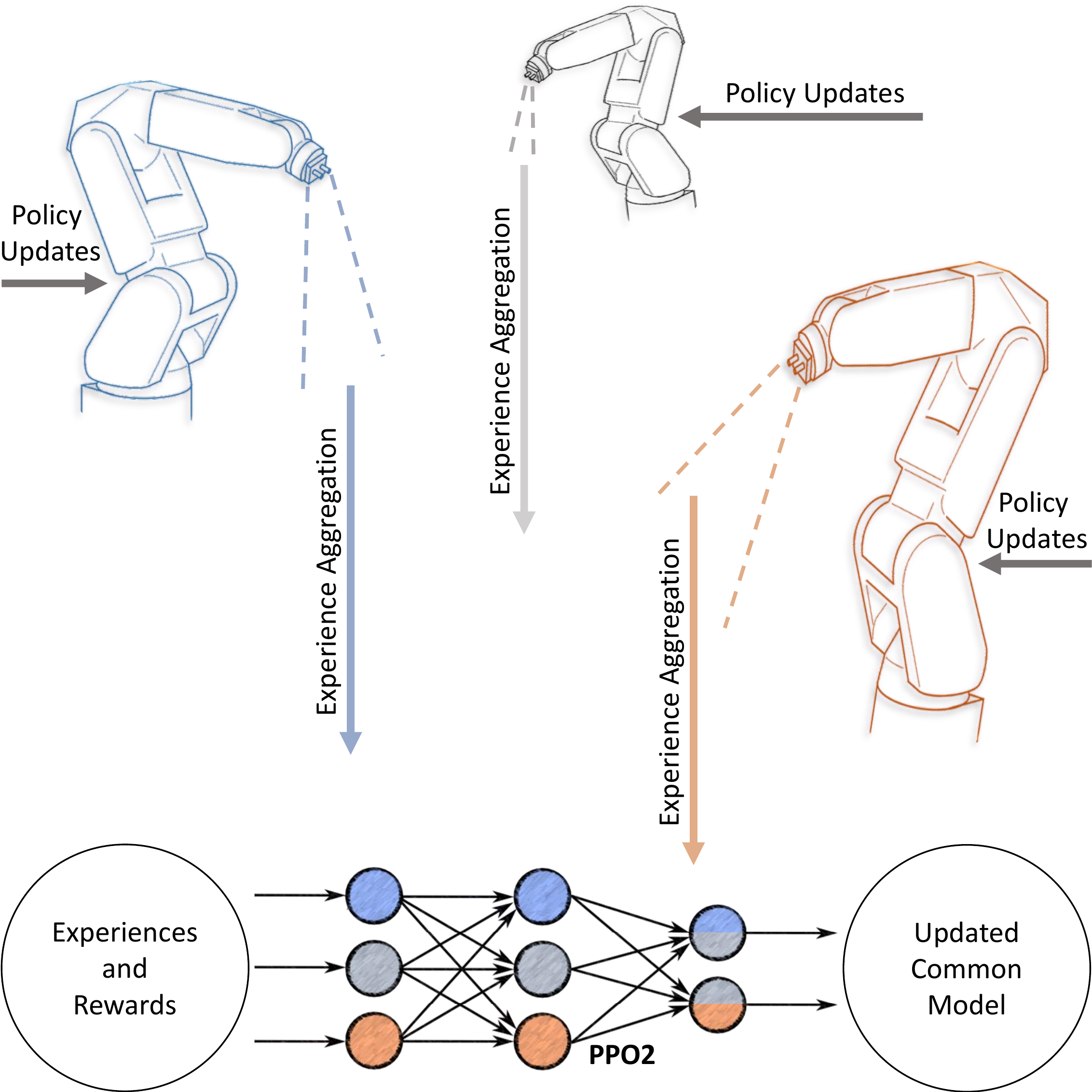}
    \caption{Conceptual view of the proposed scenario, where multiple robotic agents are collaboratively learning the same task. While the task is common, and the agents are a priori identical, we study how different alterations in the agents or their environments affects the performance of the collaborative learning process.}
    \label{fig:concept}
    \vspace{-1em}
\end{figure}

Reinforcement learning applied to multi-agent systems has two dimensions: DRL algorithms that model policies for multi-agent control and interaction, and DRL approaches that rely on multiple agents to parallelize the learning process or explore a wider variety of experiences. Within the former category, we can find examples of DRL for formation control~\cite{conde2017time}, obstacle and collision avoidance~\cite{chen2017decentralized, long2018towards}, collaborative assembly~\cite{schwung2017application}, or cooperative multi-agent control in general~\cite{gupta2017cooperative}. In the latter category, most existing approaches refer to the utilization of multiple agents to learn in parallel, but from the point of view of a multi-process or multi-threaded application~\cite{mnih2016asynchronous}. We are interested in works exploring the possibilities of using multiple robotic agents that collaborate on learning the same task. This has been identified as one of the future paradigms with 5G-and-beyond connectivity and edge computing~\cite{queralta2020enhancing, queralta2020blockchain}. For instance, in~\cite{gu2017deep} an asynchronous method for off-policy updates between robots was presented. Other works also consider network conditions and propose frameworks for multi-agent collaborative DRL over imperfect network channels~\cite{yu2020multi}. This type of scenario is illustrated in Fig.~\ref{fig:concept}, where three robotic arms are collaboratively learning the same task and sharing their experiences to update a common policy. Hereinafter, we refer to these types of scenarios as multi-agent or multi-robot collaborative RL tasks, where multiple agents collaborate to learn the same task but might be exposed to different environments, or work under different conditions.

Among the multiple challenges in DRL, recent years have seen a growing research interest in closing the simulation-to-reality gap~\cite{matas2018sim, balaji2019deepracer}, and on the design and development of robust algorithms with resilience against adversarial conditions~\cite{behzadan2017vulnerability, gleave2019adversarial, wang2020reinforcement}. This latter topic is also of paramount relevance in distributed or multi-agent DRL, where adversarial agents can hinder the collaborative learning process~\cite{song2018multi}. When multiple agents are learning a collaborative or coordinated behavior, byzantine agents can significantly reduce the performance of the system as a whole.

We aim at studying how adversarial conditions can help to bridge the simulation-to-reality gap. In~\cite{matas2018sim} and~\cite{balaji2019deepracer}, the authors analyze perturbances in the rewards towards the applicability of DRL in real-world applications. In~\cite{matas2018sim}, the focus is on learning how to manipulate deformable objects, with agents trained in a simulation environment but directly deployable in the real-world. In~\cite{arndt2019meta}, the authors present a meta-learning approach for domain adaption in simulation-to-reality transfers. Our objective in this paper is not to design a specific sim-to-real method for a given algorithm or task, but instead to analyze the performance of collaborative multi-robot DRL in the presence of disturbances in the environment as a step towards more effective sim-to-real transfers where real noises, errors or perturbances are accounted for also in the simulation environment. This includes variability in the operation of the robots, as robots might be operating in slightly different environments, or operate in different ways under the same environment. In particular, we are interested in studying how exposing multiple collaborative robots to different environments from the point of view of sensing and actuation can affect the joint learning effort.

In this paper, therefore, we focus on introducing perturbances inspired by real-world cases in a multi-agent DRL simulation. We expose different subsets of agents to slightly modified environments and study how different types of disturbances affect the collaborative learning process and the ability of the multi-robot system to converge to a common policy. The main contribution of this paper is the analysis of how input and output disturbances affect a collaborative deep reinforcement learning process with multiple robot arms. In particular, we simulate real-world perturbations that can occur on robotic arms, from the sensing and actuation perspectives. This is, to the best of our knowledge, the first study to consider the evaluation of both sensing and actuation disturbances in a multi-robot collaborative learning scenario, with different robots being exposed to different environments.

The remainder of this document is organized as follows. In Section~\ref{sec:related} we review the literature in distributed RL, adversarial RL, and robust multi-agent RL in the presence of byzantine agents. Then, Section~\ref{sec:methodology} introduces the DRL algorithm, and the methodology and simulation environment utilized in our experiments. The agent training methods and environment disturbances introduced to emulate real-world operational variability, together with the simulations results, are presented in Section~\ref{sec:results}. Section~\ref{sec:conclusion} concludes the work. 

\section{Related Works}
\label{sec:related}

In this work, we study adversarial conditions in a simulation environment to emulate real-world conditions in terms of variability of the environment across a set of multiple agents collaborating in learning the same task. With most of the literature in simulation-to-reality transfer aiming at specific applications or adaptation to different environments~\cite{balaji2019deepracer, matas2018sim, arndt2019meta}, in this section we focus instead on previous works analyzing the effect of adversarial of byzantine effects in multi-agent reinforcement learning, as well as considering other perturbations in the environment to better emulate real-world conditions. The literature in adversarial conditions for collaborative multi-agent learning is, nonetheless, sparse.

Adversarial RL has been a topic of extensive study over the past years. Multiple deep learning algorithms have been shown to be vulnerable when subject to adversarial input perturbations, being able to induce certain policies~\cite{behzadan2017vulnerability}. This is a general problem of reinforcement learning that affects different types of algorithms and scenarios. In multi-agent environments, the ability of an attacker to create adversarial observations increases significantly~\cite{gleave2019adversarial}. A comprehensive survey on the main challenges and potential solutions for adversarial attacks on DRL is available in~\cite{ilahi2020challenges}. The authors classify attacks in four categories: attacks targeting (i) rewards, (ii) policies, (iii) observations, and (iv) the environment. Among these, those targeting observations and the environment are the most relevant within the scope of this survey. In most of these cases, however, the literature only considers single-agent learning (or multiple agents being affected in the same way). Moreover, previous works focus on malicious perturbations aimed at decreasing the performance of the learning agent. In this paper, nonetheless, we induce perturbations that are inspired by real-world issues including changes in accuracy or calibration errors.

Other authors have explored the effects of having noisy rewards in RL. In this direction, Wang et al. presented an analysis of perturbed rewards for different RL algorithms, including PPO but also DQN and DDPG, among others~\cite{wang2020reinforcement}. Compared to their approach, we also consider perturbances on the RL process but focus on those that model real-world noises and errors. Moreover, we specifically put an emphasis on multi-robot collaborative learning, and consider situations in which the perturbances that affect different robots are also different. We also focus on the PPO algorithm as the state-of-the-art in three-dimensional locomotion. In fact, PPO has been identified as one of the most robust approaches against reward perturbances in~\cite{wang2020reinforcement}. Also within the study of noisy rewards, a method to improve performance in such scenarios is proposed in~\cite{kumar2019enhancing}.

In general, we see a gap in the literature in the study of noisy or perturbed environments that do not affect equally across multiple agents collaborating towards learning the same task. This paper thus tries to address this issue with an initial assessment of how perturbations in the environment influencing a subset of agents affect a global common model where experiences from different agents are aggregated.

\section{Methodology}
\label{sec:methodology}

In this section, we define our problem of distributed reinforcement learning with a subset of perturbed agents, as well as the simulation environment and modifications applied to it.

\subsection{Multi-agent RL}


In multi-agent reinforcement learning, approaches can be roughly divided into two parallel modes, asynchronous and synchronous. A3C (Asynchronous Advantage Actor-Critic)\cite{mnih2016asynchronous} is one of the most widely adopted methods for multi-agent reinforcement learning, representing the asynchronous paradigm. A3C consists of multiple independent agents with their own networks. These agents interact with different copies of the environment in parallel and update a global network periodically and asynchronously.
After each update, the agents reset their own weights to those of the global network and then resume their independent exploration.
Because some of the agents will be exploring the environment with an older version of the network weights, A3C results in relatively suboptimal use of computational resources as well as more noisy updates. An alternative is A2C (Advantage Actor-Critic), which utilizes synchronous parallel mode. In this case, there are only two networks in the system. One is used by all agents equally to interact with the environment in parallel, and outputs a batch of experiences. With this data, the second network is trained and updates the former network.


In this paper, we utilize a synchronous multi-agent reinforcement learning algorithm: proximal policy optimization (PPO). PPO and has been adopted as the default method of OpenAI owing to its excellent performance. The PPO algorithm takes advantage of the A2C ideas in terms of having multiple workers, and gradient policy ideas from TRPO (Trust Region Policy Optimization) to improve the actor performance by utilizing a trust region. PPO seeks to find a balance between the ease of implementation, sample complexity, and ease of adjustment, trying to update at each step to minimize the cost function while assuring that the new policies are not far from last policies. The scheme follows these steps:
\begin{enumerate}[wide, labelwidth=!, labelindent=0pt]
    \item Set the initial policy parameters $\theta^{0}$.
    \item In each iteration, use $\theta^{k}$ to interact with the environment, collect experience data (a tuple of state and action $\{s_{t},a_{t}\}$), and compute their advantage $A^{\theta^{k}}(s_{t},a_{t})$~\cite{mnih2016asynchronous}.
    \item Find the optimal $\theta$ by optimizing $J_{PPO}(\theta)$:
    \begin{equation}
    J_{PPO}^{\theta^{k}}(\theta)=J^{\theta^{k}}(\theta)-\beta\\KL\left(\theta,\theta^{k}\right)
    \end{equation}   
    where $\beta$ is a hyperparameter and will be adapted according to the value of $KL$. $J^{\theta^{k}}(\theta)$ is calculated by:
    \begin{equation}
    J^{\theta^{k}}(\theta)\approx \sum_{(s_{t},a_{t})}\dfrac{p_{\theta}(a_{t}\vert s_{t})}{p_{\theta^{k}}(a_{t}\vert s_{t})}A^{\theta^{k}}\left(s_{t},a_{t}\right)
    \end{equation}  
    where $p_{\theta^{k}}\left(a_{t}\vert s_{t}\right)$ is the probability of $(s_{t},a_{t})$ under $\theta^{k}$.

\end{enumerate}





\subsection{Simulation Environment}

Our simulation environment is built based on top one of the Bullet physics simulators, specifically the PyBullet Kuka arm for grasping~\cite{coumans2016pybullet}. In order to simplify the training of our RL algorithm, we modify the original grasping task into a reaching task, which allows us to focus on observing the effect of adversarial agents in training distributed reinforcement learning algorithms, rather on optimizing the training itself. 

\begin{figure}
    \centering
    \includegraphics[width=0.36\textwidth]{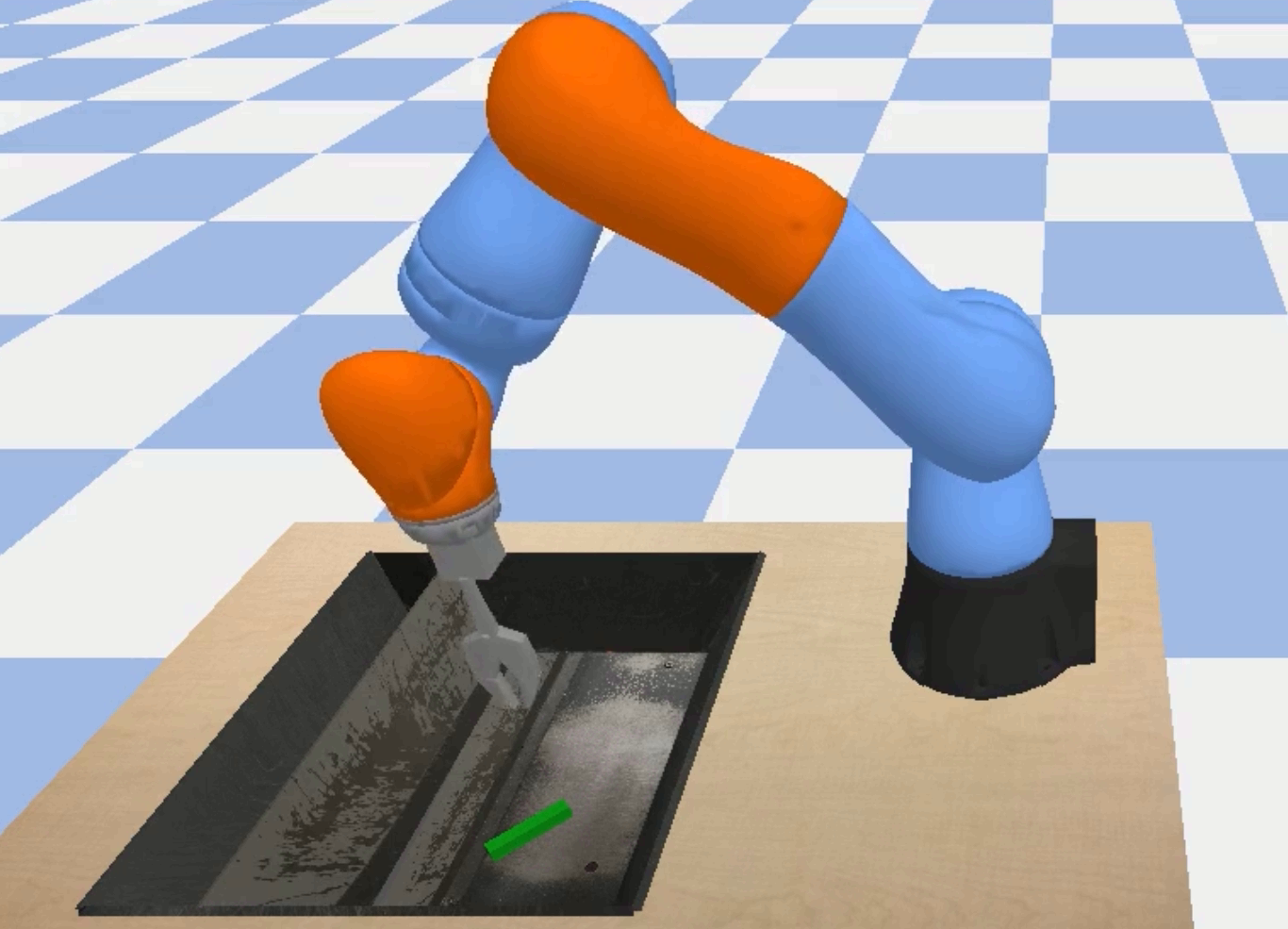}
    \caption{Kuka arm reaching environment based on Bullet simulator.}
    \label{fig:kuka_arm_env}
\end{figure}

The simulation environment is shown in Figure~\ref{fig:kuka_arm_env}. The robotic arm in this environment attempts to reach the object in the bin. It takes the Cartesian coordinates of the gripper and the relative position of the object as input instead of the on-shoulder camera observation. This input can be seen as a list with nine elements:
\begin{equation}
    Input=[x_{g},y_{g},z_{g},yaw_{g},pit_{g},rol_{g},x_{og},y_{og},rol_{og}]
\end{equation}  
where $x_{g},y_{g},z_{g}$ denote the Cartesian coordinates of the center of the gripper, and $yaw_{g},pit_{g},rol_{g}$ refers to its three Euler angles, while $x_{og},y_{og},rol_{og}$ indicate the relative $x$, $y$ position and the roll angle of the object in the gripper space.


Our RL algorithm receives the input and then gives a Cartesian displacement:
\begin{equation}
    Output=[dx, dy, dz, d\phi]
\end{equation}
in which $\phi$ is the rotation angle of the wrist around the $z$-axis. An inverse kinematics method is then employed to calculate the real motor control values of the joints. Note that all the units used for the position are in meters, and the angles are in radians. This environment with our training code is now open-source on Github\footnote{https://github.com/TIERS/NoisyKukaReacher}.

\subsection{Collaborative Learning under Real-World Perturbations}

In real robots, some of the most characteristic sources of perturbations within a homogeneous multi-robot team come from the calibration of the robots in terms of sensing and actuation. In this paper, we thus study how these two types of input (sensing) and output (actuation) perturbations affect a collaborative learning process:

\emph{Input perturbations}: we consider both fixed and variable errors in the input to the network regarding the position of the object to be reached. This emulates the error that might result from identifying the position of the object from a camera or another sensor on the robot arm. The fixed noise represents, for instance, installation or calibration errors on the position of the camera, which might have an offset in position or orientation. Variable errors, on the other hand, try to emulate the sensing errors that come, for example, from the vibration of the arm or local odometry errors describing its orientation and position.

\emph{Output perturbations}: we simulate both fixed and variable perturbations in the actuation of the robotic arm, emulating calibration errors (e.g., a constant offset in one direction), or changes in accuracy or repeatability across different robots. 

Through multiple simulations, we study how these types of perturbations affect the collaborative learning effort when they are not common across the entire set of agents. 







\begin{figure*}
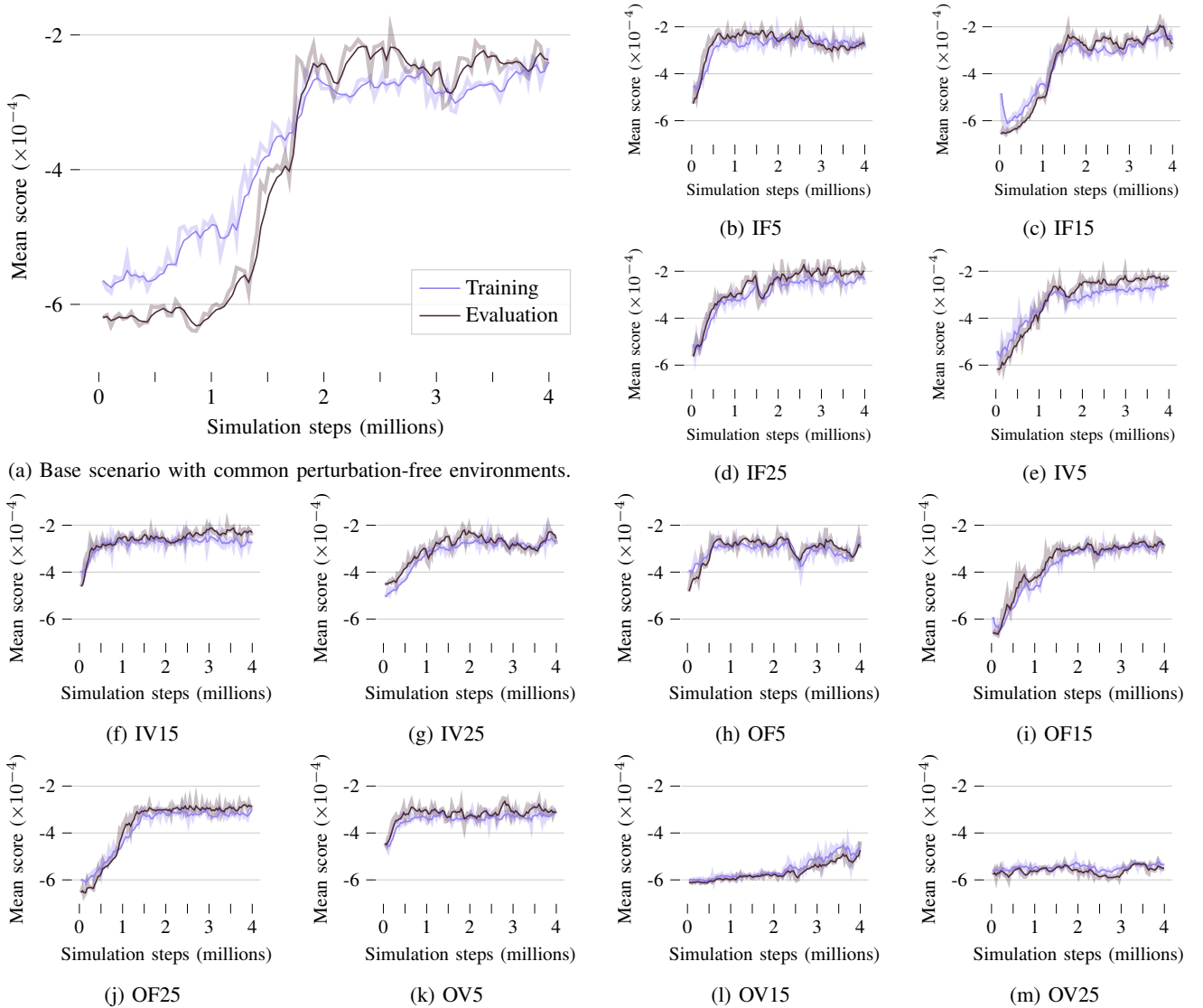

    \centering
    \begin{minipage}{0.49\textwidth}
        \begin{subfigure}[t]{0.99\textwidth}
            \centering
            \setlength{\figureheight}{0.8\textwidth}
            \setlength{\figurewidth}{\textwidth}
            \small{\input{fig_latex_v2/20200628T124652}}
            \caption{Base scenario with common perturbation-free environments.}
            \label{fig:base}
        \end{subfigure}
    \end{minipage}
    \begin{minipage}{0.49\textwidth}
        \begin{subfigure}[t]{0.49\textwidth}
            \centering
            \setlength{\figureheight}{0.8\textwidth}
            \setlength{\figurewidth}{\textwidth}
            \scriptsize{\input{fig_latex_v2/20200629T211808}}
            \caption{IF5}
            \label{fig:IF5}
        \end{subfigure}
        \begin{subfigure}[t]{0.49\textwidth}
            \centering
            \setlength{\figureheight}{0.8\textwidth}
            \setlength{\figurewidth}{\textwidth}
            \scriptsize{\input{fig_latex_v2/20200629T183036}}
            \caption{IF15}
            \label{fig:IF15}
        \end{subfigure}
        \begin{subfigure}[t]{0.49\textwidth}
            \centering
            \setlength{\figureheight}{0.8\textwidth}
            \setlength{\figurewidth}{\textwidth}
            \scriptsize{\input{fig_latex_v2/20200629T202018}}
            \caption{IF25}
            \label{fig:IF25}
        \end{subfigure}
        \begin{subfigure}[t]{0.49\textwidth}
            \centering
            \setlength{\figureheight}{0.8\textwidth}
            \setlength{\figurewidth}{\textwidth}
            \scriptsize{\input{fig_latex_v2/20200628T173606}}
            \caption{IV5}
            \label{fig:IV5}
        \end{subfigure}
    \end{minipage}
    \begin{subfigure}[t]{0.24\textwidth}
        \centering
        \setlength{\figureheight}{0.8\textwidth}
        \setlength{\figurewidth}{\textwidth}
        \footnotesize{\input{fig_latex_v3/20200630T111226}}
        \caption{IV15}
        \label{fig:IV15}
    \end{subfigure}
    \begin{subfigure}[t]{0.24\textwidth}
        \centering
        \setlength{\figureheight}{0.8\textwidth}
        \setlength{\figurewidth}{\textwidth}
        \footnotesize{\input{fig_latex_v2/20200628T191601}}
        \caption{IV25}
        \label{fig:IV25}
    \end{subfigure}
    \begin{subfigure}[t]{0.24\textwidth}
        \centering
        \setlength{\figureheight}{0.8\textwidth}
        \setlength{\figurewidth}{\textwidth}
        \footnotesize{\input{fig_latex_v3/20200630T120538}}
        \caption{OF5}
        \label{fig:OF5}
    \end{subfigure}
    \begin{subfigure}[t]{0.24\textwidth}
        \centering
        \setlength{\figureheight}{0.8\textwidth}
        \setlength{\figurewidth}{\textwidth}
        \footnotesize{\input{fig_latex_v2/20200629T165257}}
        \caption{OF15}
        \label{fig:OF15}
    \end{subfigure}
    \begin{subfigure}[t]{0.24\textwidth}
        \centering
        \setlength{\figureheight}{0.8\textwidth}
        \setlength{\figurewidth}{\textwidth}
        \footnotesize{\input{fig_latex_v2/20200628T222909}}
        \caption{OF25}
        \label{fig:OF25}
    \end{subfigure}
    \begin{subfigure}[t]{0.24\textwidth}
        \centering
        \setlength{\figureheight}{0.8\textwidth}
        \setlength{\figurewidth}{\textwidth}
        \footnotesize{\input{fig_latex_v3/20200630T131604}}
        \caption{OV5}
        \label{fig:OV5}
    \end{subfigure}
    \begin{subfigure}[t]{0.24\textwidth}
        \centering
        \setlength{\figureheight}{0.8\textwidth}
        \setlength{\figurewidth}{\textwidth}
        \footnotesize{\input{fig_latex_v2/20200629T060207}}
        \caption{OV15}
        \label{fig:OV15}
    \end{subfigure}
    \begin{subfigure}[t]{0.24\textwidth}
        \centering
        \setlength{\figureheight}{0.8\textwidth}
        \setlength{\figurewidth}{\textwidth}
        \footnotesize{\input{fig_latex_v2/20200629T070507}}
        \caption{OV25}
        \label{fig:OV25}
    \end{subfigure}
    \caption{Simulation results where we show the training in perturbance-free environment, and 12 cases where we analyze the effect of a modified environment (fixed and variable perturbances on both sensing and actuation) on 5, 15 and 25 agents. The total number of agents is 30 in all cases. The legend is common across all graphs and has been omitted in subfigures (b) through (m) to improve readability.}
    \label{fig:main_results}
\end{figure*}

\section{Experimentation and Results}
\label{sec:results}

In this section, we describe the training parameters utilized through our simulations, and the ways in which the environments have been modified to introduce disturbances in both sensors and actuators. We then present the results of multiple simulations where different numbers of agents have been trained in different environments but treated equally from the point of view of the collaborative learning process.

\subsection{Training Method}

The maximal number of steps in one episode is set to 1200, and the maximum number of steps for the whole training process is set to 4 million. If the gripper contacts the object or approaches it at a very small distance (0.008\,m), the episode will be terminated. The final score for this episode is thus calculated by summing all the rewards obtained in all the steps until termination.

The initial reward is set as -1000 for each step if the distance is larger than 1\,m. However, if the distance between the finger on the gripper and the object is smaller than 1\,m, the reward is computed as $reward_{raw} = -10\cdot distance$. Moreover, we also add the cost of each step, in order to encourage the gripper to approach the target as soon as possible. The cost of each step is set as 1. Therefore, the final reward for this step is hence: $ reward_{final} = -10\cdot distance - 1$, where the distance is given in meters. If the gripper finally contacts its target or approach it in a threshold, we give it a significantly larger reward (1000) to help the model learn faster and clearly.

In total, in our simulations, we utilize 30 agents parallelized on the GPU processes to produce experience data based on a vectorized environment. Therefore, these agents can represent a multi-robot system learning a collaborative RL task. We give different settings on individual environments to manually simulate the possible perturbations that robots find in real-world scenarios.

\subsection{Calibration and Accuracy Noises}

To emulate the practical noises and errors that robots could encounter when training an RL algorithm, we consider the following four types of perturbations, for each of which we generate a different environment to expose a variable number of robots to: fixed input errors on all the nine elements by $0.005\,m$, uniformly distributed sensing errors in the interval $[0.005\,m, 0.01\,m]$, fixed output errors modifying the gripper actuators by an offset of $0.005\,m$ on the $x$ axis, and uniformly distributed output errors in the interval $[0.005\,m, 0.01\,m]$. It should be noted that the uniform distributed errors on input and output could be different in each step, which can be regarded as inaccurate sensing errors, or reduced repeatability in the actuation of real robots.

Moreover, in order to further analyze how more extreme cases affect the collaborative learning process, we also consider fixed disturbances on larger magnitude (0.015\,m on all the values for the sensing error and 0.015\,m on the x-axis for the actuation error) as well as scenarios where the noise is different for each of the agents exposed to the modified environment (in the interval 0.005\,m to 0.025\,m for 25 agents).

\subsection{Simulation Results}

Figure~\ref{fig:main_results} shows the results of our simulations. The notation describing each subfigure is as follows: \{I,O\}: representing the input (sensing) and output (actuation) perturbances, \{F,V\}: representing fixed and variable perturbances, and \{5, 15, 25\}: representing the number of agents exposed to the modified environment where the perturbances occur.

\begin{figure*}
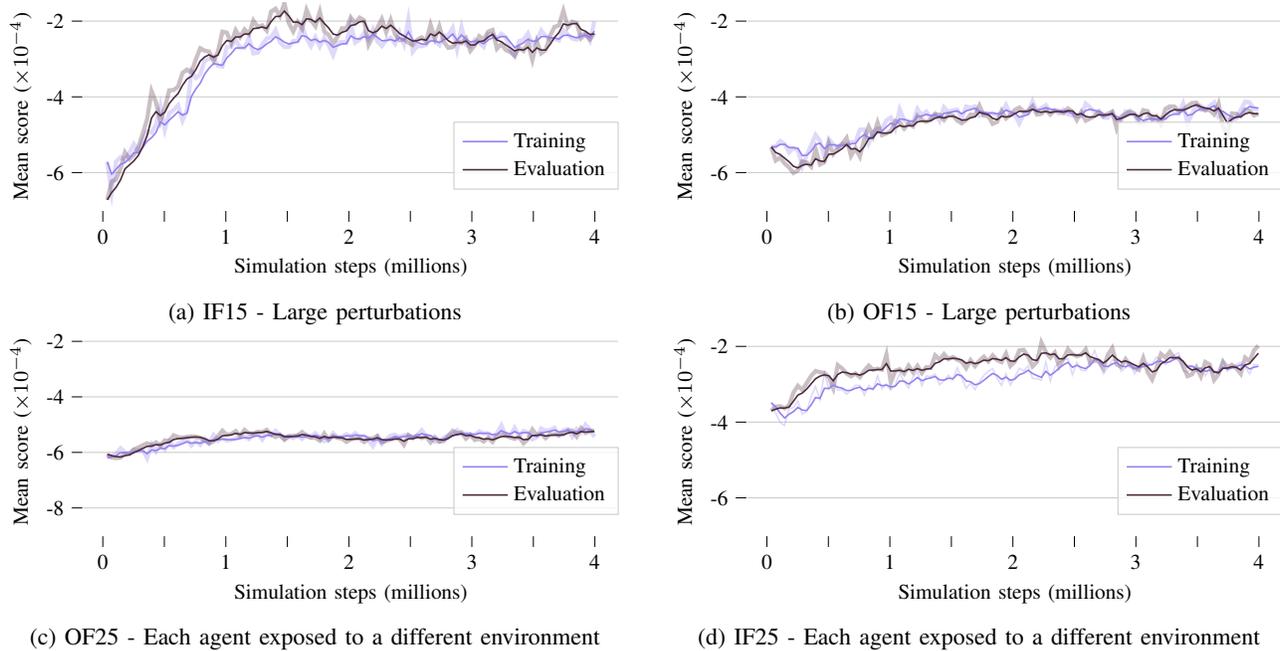

    \centering
    \begin{subfigure}[t]{0.48\textwidth}
        \centering
        \setlength{\figureheight}{0.5\textwidth}
        \setlength{\figurewidth}{\textwidth}
        \footnotesize{\input{fig_latex_v2/20200629T135403}}
        \caption{IF15 - Large perturbations}
        \label{fig:IF15BigNoise}
    \end{subfigure}
    \begin{subfigure}[t]{0.48\textwidth}
        \centering
        \setlength{\figureheight}{0.5\textwidth}
        \setlength{\figurewidth}{\textwidth}
        \footnotesize{\input{fig_latex_v2/20200629T145641}}
        \caption{OF15 - Large perturbations}
        \label{fig:OF15BigNoise}
    \end{subfigure}
    \begin{subfigure}[t]{0.48\textwidth}
        \centering
        \setlength{\figureheight}{0.5\textwidth}
        \setlength{\figurewidth}{\textwidth}
        \footnotesize{\input{fig_latex_v2/20200629T101439}}
        \caption{OF25 - Each agent exposed to a different environment}
        \label{fig:OFMN25}
    \end{subfigure}
    \begin{subfigure}[t]{0.48\textwidth}
        \centering
        \setlength{\figureheight}{0.5\textwidth}
        \setlength{\figurewidth}{\textwidth}
        \footnotesize{\input{fig_latex_v3/20200630T161240}}
        \caption{IF25 - Each agent exposed to a different environment}
        \label{fig:IFMN25}
    \end{subfigure}
    \caption{Simulation results with an extra 4 scenarios analyzed: two of them where we consider perturbations of larger magnitude, and two more where we consider that each of the agents in a modified environment is affected differently.}
    \label{fig:more_results}
\end{figure*}

Comparing perturbations in the sensors versus perturbations in the actuators, we see an overall more robust performance against adversarial elements in the sensing part. In Figures~\ref{fig:IF5} to~\ref{fig:IF25}, we see that the network always converges and we only see more unstable behaviour when there is a large fraction of agents suffering of variable sensing errors (50\% and 83\%). When we compare the effect of constant or fixed perturbations against variable ones, we notice that variable perturbations induce less stable convergence. This can be to some extent explained by the fact that there are no large subsets of agents being exposed to a common environment.

For small fixed perturbations affecting actuation (output disturbances), we have seen that the agents are able to converge towards a working policy. In the cases where 5 or 25 of the agents are affected, this was expected as there is a majority (25) of agents, in both cases, that work in exactly the same way, and a small subset (5) that work in a slightly different way (but still the same within that subset). When this fixed perturbation is introduced to half of the agents, then we have two subsets of the same size operating in different ways, but again consistently across each of the subsets. In this case, we have seen that for a small magnitude in the perturbation, the agents still converge on a policy that works for both subsets. As the difference between the operation of the agents in these two subsets diverges, the performance of the system as a whole drops significantly. Nonetheless, we have observed that the case were half of the agents have a common fixed perturbation of small magnitude the system is able to converge even when the initial conditions are disadvantageous. 

In order to analyze the effect of perturbations with larger magnitude as well as fixed perturbations in both sensing and actuation that vary across the robots exposed to a modified environment, we have analyzed four more cases shown in Fig.~\ref{fig:more_results}. In Figures~\ref{fig:IF15BigNoise} and~\ref{fig:OF15BigNoise}, we analyze how perturbations with larger magnitude affect the learning process, with half of the agents being affected as the worst-case scenario. We see that the trend from the previous results is followed, with the network being able to converge to a common policy when a constant error is added to the sensors interface, but not when the disturbances affect to the actuators. Finally, Figures~\ref{fig:OFMN25} and~\ref{fig:IFMN25} show that when there are no differentiated subsets of agents with a common behaviour and the perturbations are different across a large number of agents, then the system is not able to converge.

\section{Conclusion and Future Work}\label{sec:conclusion}

Adversarial agents and closing the simulation-to-reality gap are among the key challenges preventing wider adoption of reinforcement learning in real-world applications. In this paper, we have addressed the latter one from the perspective of the former: by introducing adversarial conditions inspired by real-world perturbances to a subset of agents in a multi-robot system during a collaborative reinforcement learning process, we have been able to identify points where the robustness of distributed multi-agent DRL algorithms needs to be improved. In this paper, we have considered multiple robotic arms in a simulation environment collaborating towards learning a common policy to reach an object. In order to emulate more realistic conditions and understand how perturbances in the environment affect the learning process, we have considered variability across the agents in terms of their ability to sense and actuate accurately. We have shown how different types of disturbances in the model's input (sensing) and output (actuation) affect the robustness and ability to converge towards an effective policy. We have seen how variable perturbances have the most effect on the ability of the network to converge, while disturbances in the ability of the robots to actuate properly have had a comparatively worse effect than those in their ability to sense the position of the object accurately.

The conclusions of this work serve as a starting point towards the design and development of more robust methods able to identify and take into account these disturbances in the environment that do not occur across all robots equally. This will be the subject of our future work, as well as the study of other types or combinations of disturbances in the environment. We will also work towards modeling more accurately real-world errors for RL simulation environments.


\section*{Acknowledgements}

This work was supported by the Academy of Finland's AutoSOS project with grant number 328755.

\bibliographystyle{unsrt}
\bibliography{ref}

\begin{thebibliography}{10}

\bibitem{arulkumaran2017brief}
Kai Arulkumaran, Marc~Peter Deisenroth, Miles Brundage, and Anil~Anthony
  Bharath.
\newblock A brief survey of deep reinforcement learning.
\newblock {\em arXiv preprint arXiv:1708.05866}, 2017.

\bibitem{nguyen2020deep}
Thanh~Thi Nguyen, Ngoc~Duy Nguyen, and Saeid Nahavandi.
\newblock Deep reinforcement learning for multiagent systems: A review of
  challenges, solutions, and applications.
\newblock {\em IEEE transactions on cybernetics}, 2020.

\bibitem{mnih2016asynchronous}
Volodymyr Mnih, Adria~Puigdomenech Badia, Mehdi Mirza, Alex Graves, Timothy
  Lillicrap, Tim Harley, David Silver, and Koray Kavukcuoglu.
\newblock Asynchronous methods for deep reinforcement learning.
\newblock In {\em International conference on machine learning}, 2016.

\bibitem{rajeswaran2017learning}
Aravind Rajeswaran, Vikash Kumar, Abhishek Gupta, Giulia Vezzani, John
  Schulman, Emanuel Todorov, and Sergey Levine.
\newblock Learning complex dexterous manipulation with deep reinforcement
  learning and demonstrations.
\newblock {\em arXiv preprint arXiv:1709.10087}, 2017.

\bibitem{matas2018sim}
Jan Matas, Stephen James, and Andrew~J Davison.
\newblock Sim-to-real reinforcement learning for deformable object
  manipulation.
\newblock {\em arXiv preprint arXiv:1806.07851}, 2018.

\bibitem{conde2017time}
Ronny Conde, Jos{\'e}~Ram{\'o}n Llata, and Carlos Torre-Ferrero.
\newblock Time-varying formation controllers for unmanned aerial vehicles using
  deep reinforcement learning.
\newblock {\em arXiv preprint arXiv:1706.01384}, 2017.

\bibitem{chen2017decentralized}
Yu~Fan Chen, Miao Liu, Michael Everett, and Jonathan~P How.
\newblock Decentralized non-communicating multiagent collision avoidance with
  deep reinforcement learning.
\newblock In {\em ICRA}, pages 285--292. IEEE, 2017.

\bibitem{long2018towards}
Pinxin Long, Tingxiang Fanl, Xinyi Liao, Wenxi Liu, Hao Zhang, and Jia Pan.
\newblock Towards optimally decentralized multi-robot collision avoidance via
  deep reinforcement learning.
\newblock In {\em ICRA}. IEEE, 2018.

\bibitem{schwung2017application}
Dorothea Schwung, Fabian Csaplar, Andreas Schwung, and Steven~X Ding.
\newblock An application of reinforcement learning algorithms to industrial
  multi-robot stations for cooperative handling operation.
\newblock In {\em 15th INDIN}, pages 194--199. IEEE, 2017.

\bibitem{gupta2017cooperative}
Jayesh~K Gupta, Maxim Egorov, and Mykel Kochenderfer.
\newblock Cooperative multi-agent control using deep reinforcement learning.
\newblock In {\em AAMAS}, pages 66--83. Springer, 2017.

\bibitem{queralta2020enhancing}
Jorge Pe\~{n}a Queralta, Li~Qingqing, Zhuo Zou, and Tomi Westerlund.
\newblock Enhancing autonomy with blockchain and multi-acess edge computing in
  distributed robotic systems.
\newblock In {\em The Fifth International Conference on Fog and Mobile Edge
  Computing (FMEC). IEEE}, 2020.

\bibitem{queralta2020blockchain}
Jorge {Pe\~{n}a Queralta} and Tomi Westerlund.
\newblock Blockchain-powered collaboration in heterogeneous swarms of robots.
\newblock {\em Frontiers in Robotics and AI}, 2020.

\bibitem{gu2017deep}
Shixiang Gu, Ethan Holly, Timothy Lillicrap, and Sergey Levine.
\newblock Deep reinforcement learning for robotic manipulation with
  asynchronous off-policy updates.
\newblock In {\em ICRA}, pages 3389--3396. IEEE, 2017.

\bibitem{yu2020multi}
Yiding Yu, Soung~Chang Liew, and Taotao Wang.
\newblock Multi-agent deep reinforcement learning multiple access for
  heterogeneous wireless networks with imperfect channels.
\newblock {\em arXiv preprint arXiv:2003.11210}, 2020.

\bibitem{balaji2019deepracer}
Bharathan Balaji, Sunil Mallya, Sahika Genc, Saurabh Gupta, Leo Dirac, Vineet
  Khare, Gourav Roy, Tao Sun, Yunzhe Tao, Brian Townsend, et~al.
\newblock Deepracer: Educational autonomous racing platform for experimentation
  with sim2real reinforcement learning.
\newblock {\em arXiv:1911.01562}, 2019.

\bibitem{behzadan2017vulnerability}
Vahid Behzadan and Arslan Munir.
\newblock Vulnerability of deep reinforcement learning to policy induction
  attacks.
\newblock In {\em MLDM}. Springer, 2017.

\bibitem{gleave2019adversarial}
Adam Gleave, Michael Dennis, Cody Wild, Neel Kant, Sergey Levine, and Stuart
  Russell.
\newblock Adversarial policies: Attacking deep reinforcement learning.
\newblock {\em arXiv preprint arXiv:1905.10615}, 2019.

\bibitem{wang2020reinforcement}
Jingkang Wang, Yang Liu, and Bo~Li.
\newblock Reinforcement learning with perturbed rewards.
\newblock In {\em AAAI}, pages 6202--6209, 2020.

\bibitem{song2018multi}
Jiaming Song, Hongyu Ren, Dorsa Sadigh, and Stefano Ermon.
\newblock Multi-agent generative adversarial imitation learning.
\newblock In {\em Advances in neural information processing systems}, pages
  7461--7472, 2018.

\bibitem{arndt2019meta}
Karol Arndt, Murtaza Hazara, Ali Ghadirzadeh, and Ville Kyrki.
\newblock Meta reinforcement learning for sim-to-real domain adaptation.
\newblock {\em arXiv preprint arXiv:1909.12906}, 2019.

\bibitem{ilahi2020challenges}
Inaam Ilahi, Muhammad Usama, Junaid Qadir, Muhammad~Umar Janjua, Ala Al-Fuqaha,
  Dinh~Thai Hoang, and Dusit Niyato.
\newblock Challenges and countermeasures for adversarial attacks on deep
  reinforcement learning.
\newblock {\em arXiv preprint arXiv:2001.09684}, 2020.

\bibitem{kumar2019enhancing}
Aashish Kumar et~al.
\newblock {\em Enhancing performance of reinforcement learning models in the
  presence of noisy rewards}.
\newblock PhD thesis, 2019.

\bibitem{coumans2016pybullet}
Erwin Coumans and Yunfei Bai.
\newblock Pybullet, a python module for physics simulation for games, robotics
  and machine learning.
\newblock 2016.

\end{thebibliography}


\end{document}